\documentclass{bmvc2k}

\usepackage{booktabs}
\usepackage{color}
\usepackage{graphicx} 
\usepackage{amssymb}
\usepackage{bbm}

\title{Boundary Guided Context Aggregation for Semantic Segmentation}

\addauthor{Haoxiang Ma$^{1,2}$}{mahaoxiang822@buaa.edu.cn}{1}
\addauthor{Hongyu Yang$^{\dag 3}$}{hongyuyang@buaa.edu.cn}{2}
\addauthor{Di Huang$^{1,2}$}{dhuang@buaa.edu.cn}{3}
\addinstitution{
 State Key Laboratory of Software Development Environment$^1$\\
 Beihang University, Beijing, China
}
\addinstitution{
 School of Computer Science and Engineering$^2$\\
 Beihang University, Beijing, China
}
\addinstitution{
 Institute of Artificial Intelligence$^3$\\
 Beihang University, Beijing, China
}

\runninghead{Ma, Yang, Huang}{Boundary Guided Context Aggregation}

\begin{document}

\maketitle

\begin{abstract}
   The recent studies on semantic segmentation are starting to notice the significance of the boundary information, where most approaches see boundaries as the supplement of semantic details. However, simply combing boundaries and the mainstream features cannot ensure a holistic improvement of semantics modeling. In contrast to the previous studies, we exploit boundary as a significant guidance for context aggregation to promote the overall semantic understanding of an image. To this end, we propose a Boundary guided Context Aggregation Network (BCANet), where a Multi-Scale Boundary extractor (MSB) borrowing the backbone features at multiple scales is specifically designed for accurate boundary detection. Based on which, a Boundary guided Context Aggregation module (BCA) improved from Non-local network is further proposed to capture long-range dependencies between the pixels in the boundary regions and the ones inside the objects. By aggregating the context information along the boundaries, the inner pixels of the same category achieve mutual gains and therefore the intra-class consistency is enhanced. We conduct extensive experiments on the Cityscapes and ADE20K databases, and comparable results are achieved with the state-of-the-art methods, clearly demonstrating the effectiveness of the proposed one. Our code is available at {\small \url{https://github.com/mahaoxiang822/Boundary-Guided-Context-Aggregation}}.
\end{abstract}


\begin{figure*}[t]
\centering
\subfigure[]{
\centering
\begin{minipage}[t]{0.23\linewidth}
\includegraphics[width=1\linewidth]{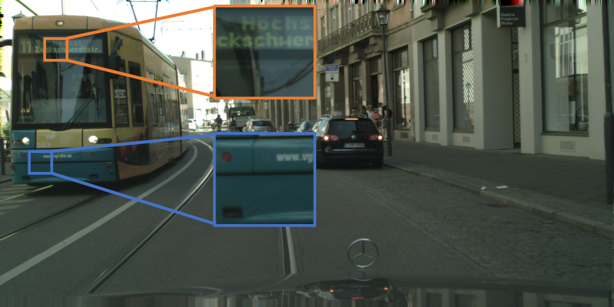}
\end{minipage}
}
\subfigure[]{
\centering
\begin{minipage}[t]{0.23\linewidth}
\includegraphics[width=1\linewidth]{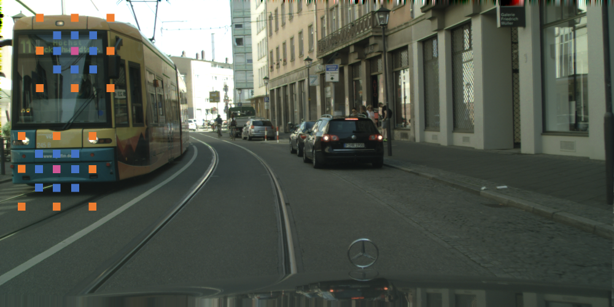}
\end{minipage}
}
\subfigure[]{
\centering
\begin{minipage}[t]{0.23\linewidth}
\includegraphics[width=1\linewidth]{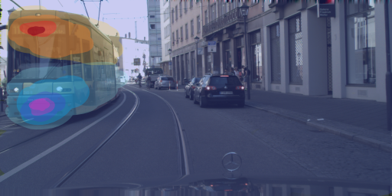}
\end{minipage}
}
\subfigure[]{
\centering
\begin{minipage}[t]{0.23\linewidth}
\includegraphics[width=1\linewidth]{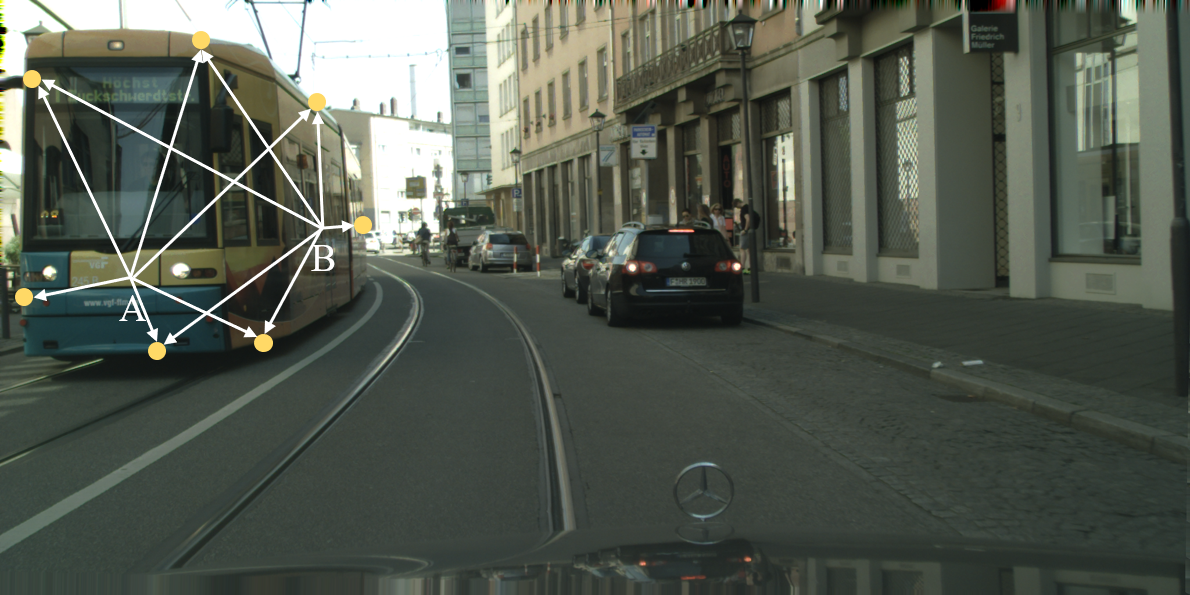}
\end{minipage}
}

\caption{Comparison of various manners for context aggregation. (a) The original input image; (b) non-adaptive context aggregation; (c) attention-based context aggregation; and (d) the proposed boundary guided context aggregation.}
\label{fig1}
\vspace{-0.5cm}
\end{figure*}

\vspace{-0.3cm}
\section{Introduction}

Semantic segmentation is a fundamental and long-standing task in computer vision, which aims to get a dense prediction of the object category for every pixel in an image. It has been extensively and actively applied to many challenging applications, \textit{e.g.}, autonomous driving \cite{cordts2016cityscapes}, image editing \cite{aksoy2018semantic}, and human-machine interaction \cite{oberweger2015hands}. 

Recently, a number of approaches based on fully convolutional networks (FCN) \cite{long2015fully} have been proposed for semantic segmentation. Due to the fixed geometric structure and the limited receptive field, however, these methods expose a common drawback in contextual information modeling. To make up for the above deficiency, a number of studies explore contextual dependencies for improved results and the existing methods can be briefly summarized into two categories. One is to use pyramid-based module which combines different scales of atrous convolution or pooling layers to enlarge the receptive field, such as Atrous Spatial Pyramid Pooling (ASPP) in Deeplab \cite{chen2017deeplab} and Pyramid Pooling Module (PPM) in PSPNet \cite{zhao2017pyramid}. Another trend is to adaptively model long-range dependencies from the aspect of channel or spatial. For example, EncNet \cite{zhang2018context} learns to weight the feature map on channels and Non-local module \cite{wang2018non} exploits the attention-based mechanism to enable one pixel to perceive all the other positions of an image. These alternatives achieve promising results on semantic segmentation. However, either aggregating contextual information in a fixed manner or with a learned response map, the existing context aggregation modules generally lack an explicit prior on the regions to aggregate that undesired inter-class dependency and intra-class ambiguity are also involved, as illustrated in Figure \ref{fig1}(b) and (c). The activated context regions are not properly regularized in the previous methods and this will inevitably impose adverse effect on the contexts. 

Considering that object boundary regions contain less mixed dependencies and they are important clues that all the inner pixels share in common, we therefore investigate a boundary constraint for context aggregation in this study, that only the dependencies between the inner pixels and boundaries are built. We notice that boundaries have been actively exploited for improved segmentation performance. In particular, the traditional post-processing modules regard boundary as a useful constraint in domain transform or global energy function, and the recent Boundary-aware Feature Propagation (BFP) module extends this idea into deep models. However, these operations are usually computationally complex and they are not easy to be integrated into an end-to-end model \cite{chen2016semantic,bertasius2016semantic,ding2019boundary}. Moreover, the methods above heavily rely on the closure of the detected boundary, otherwise the context information will be leaked into the neighboring areas and downgrades the inter-class distinction. Such fact indicates that there still leaves much space for improvement.

In contrast to the existing methods, this study directly wires boundary as the context to be aggregated. See Figure \ref{fig1}(d) for an illustration, even though the local surrounded context of the two marked pixels are quite different, their features are more likely to be close to each other once their relationships with the boundary are well established. By aggregating the semantic information along the corresponding boundary region into the inner pixels, any positions in this category harvest similar context, regardless of their conspicuous texture difference. Compared with the previous context solutions that directly model dependencies within a peripheral region or the full-image, the boundary constraint used in this study can not only exclude the undesired inter-class context relationships explicitly, but also alleviate intra-class ambiguity, e.g. an advertisement posted on a bus.

To this end, we propose a novel framework, named Boundary guided Context Aggregation network (BCANet), for image semantic segmentation. It casts boundary detection as an independent sub-task along with the mainstream feature learning, and develops an attention based mechanism for boundary guided context aggregation. Specifically, the Multi-Scale Boundary extractor (MSB) is introduced to predict the binary boundary of an image, which borrows semantic features from multiple stages of the backbone as its input and feeds back the semantic embedded boundary features to the mainstream for context aggregation. The Boundary guided Context Aggregation module (BCA) modified from the Non-local network calculates the attention map between the boundary features and the mainstream semantics, so that boundaries can be seen as a mutual guidance for aggregating the context information, which enables the pixels of the same class achieve similar gains. Extensive evaluations validate that the proposed method performs favorably against the current state-of-the-art context aggregation approaches.

\vspace{-0.4cm}
\section{Related Work}
\vspace{-0.2cm}
\subsection{Semantic Segmentation}

In the recent years, the Convolutional Neural Networks (CNNs) based methods are dominating the field of semantic segmentation. Fully Convolutional Network (FCN) \cite{long2015fully} opens a precedent for the application of CNNs in semantic segmentation by replacing the fully connected layers with convolution layers.
More recently, a number of studies focus on capturing richer context information to augment the feature representation, where the multi-scale pyramid modules promote the performance a lot. For example, Zhao \emph{et al.} \cite{zhao2017pyramid} investigate PPM in PSPNet to aggregate contextual information at
different scales, and Chen \emph{et al.} \cite{chen2017deeplab} introduce the ASPP module.
Besides, to overcome the limitation of fixed size of the kernel, several global pooling based methods \cite{zhang2018context,yu2018learning} and graph convolutional based methods \cite{zhang2019dual,zhang2020dynamic} are introduced to perceive the global context. To overcome the limitation of labeled training data, Chen \emph{et al.} \cite{chen2020naive} develop a semi-supervised method and they achieve the state-of-the-art results. Our study also leverages the context information, but we see boundary as a more effective and explicit constraint.

\vspace{-0.2cm}
\subsection{Boundary in Semantic Segmentation}
As an essential element of image, boundary has been paid much attention in computer vision. In the early research on FCN-based semantic segmentation, \cite{chen2016semantic,bertasius2016semantic,chen2017deeplab} use boundaries for post-processing to refine the result at the end of the network.
Recently, several methods are starting to explicitly model boundary detection as an independent subtask in parallel with semantic segmentation for sharper results. Takikawa \emph{et al.} \cite{takikawa2019gated} and Zhen \emph{et al.} \cite{zhen2020joint} specially design a boundary stream and couple the two tasks of boundary and semantics modeling. Li \emph{et al.} \cite{li2020improving} point out that the object boundary and body parts correspond to the high frequency and low frequency information of an image, respectively, based on which they decouple the body and edge with diverse supervisions. Yuan \emph{et al.} \cite{yuan2020segfix} propose a model-agnostic method for boundary refinement. Different from the existing studies, our work explicitly exploits the boundary information for context aggregation which will further enhance the semantic representations, rather than just simply combining them. Ding \emph{et al.} \cite{ding2019boundary} investigate a Boundary-aware Feature Propagation module (BFP) to propagate information among the inner pixels of an object, which shares a common concept with our work. However, the full feature map will become over-smooth in BFP once the detected boundary is not closured, thus it greatly downgrades the inter-class distinction.

\subsection{Attention Mechanism in Semantic Segmentation}

Attention mechanism has been actively exploited in deep neural networks. For semantic segmentation, Chen \emph{et al.} \cite{chen2016attention} present a pioneering work that utilizes attention to reweight the multi-scale features. Inspired by the use of self-attention in machine translation \cite{vaswani2017attention}, Wang \emph{et al.} \cite{wang2018non} propose the Non-local module to capture global dependencies by calculating the correlation matrix of pixels at all positions and apply it to generate new feature representations. Based on Non-local module, a number of methods  \cite{fu2019dual,huang2019ccnet,yuan2020object,yu2020context,yin2020disentangled} are proposed for more accurate semantic segmentation.
Our BCA module is also inspired by attention mechanism, we present a different but more effective network, where it mainly focuses on the positions in boundary regions. This solution is more powerful in reinforcing the intra-class correlation than the previous methods when intra-class ambiguity occurs.

\vspace{-0.4cm}
\section{Methodology}
\vspace{-0.2cm}

\begin{figure*}[t]
\centering
\includegraphics[width=0.98\linewidth]{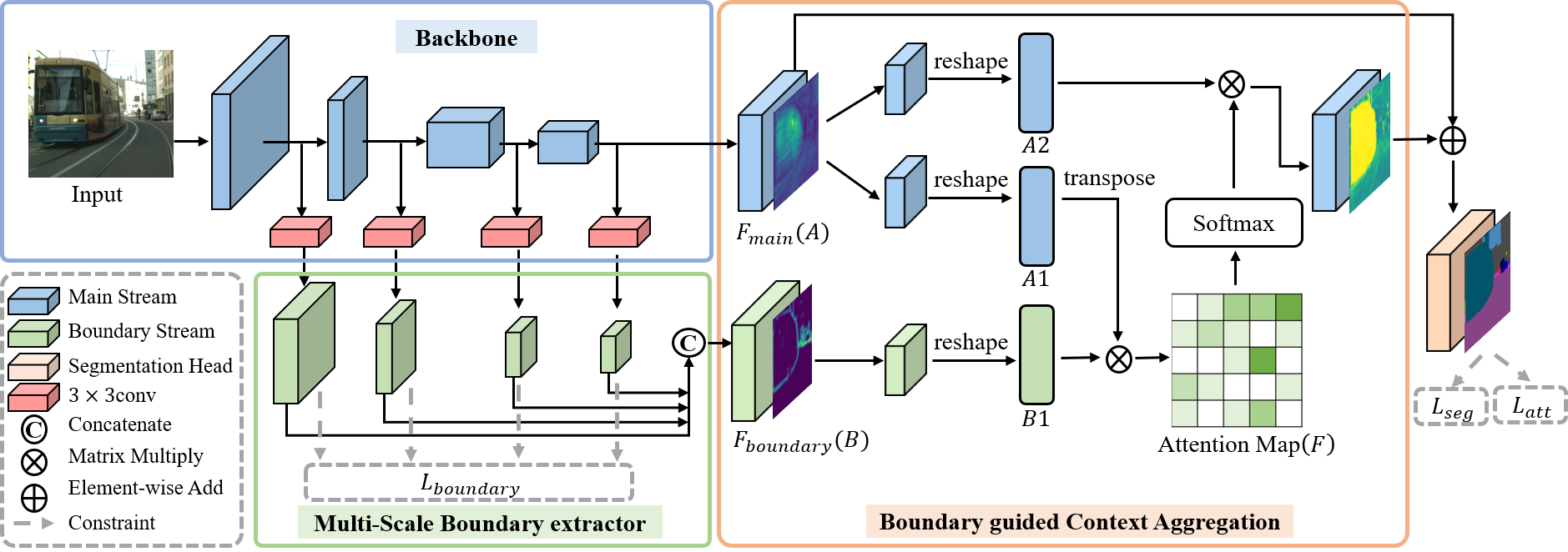}
\caption{An overview of the proposed Boundary guided Context Aggregation Network (BCANet). With the semantic embedded boundary features generated by the Multi-Scale Boundary module (MSB), the Boundary guided Context Aggregation module (BCA) employs a cross-stream attention for more accurate context aggregation.}
\label{fig2}
\vspace{-0.5cm}
\end{figure*}

The proposed method explicitly models boundaries as the context, ensuring that inner pixels within the same object achieve similar gains after context aggregation. See Figure \ref{fig2} for an overview. Specifically, the method employs the residual networks \cite{he2016deep} as its backbone, where the vanilla convolutions are replaced by the dilated ones to enlarge the receptive field. Furthermore, we introduce the Multi-Scale Boundary extractor (MSB) to model boundary prediction as an independent sub-task along with the mainstream feature learning. The features from each stage of the backbone with different scales are concatenated together to capture boundary information with diverse semantics. With the semantic embedded boundary features generated by MSB, the Boundary guided Context Aggregation module (BCA) is carefully designed to perform semantic context aggregation. We build relationships between each pixel in the semantic feature maps and the boundary features to generate an attention map where the object boundary regions of the same category are highly activated. By aggregating the context information along the boundaries, pixels of the same category can achieve similar gains, thus enhancing the semantic consistency. We detail the method subsequently.

\vspace{-0.2cm}
\subsection{Multi-Scale Boundary (MSB) extractor}

Considering that the boundary features are used to guide the process of semantic context aggregation and they interact with the mainstream features, MSB thus directly borrows the intermediate representations from the backbone as its input. We add multiple connections between the mainstream and the boundary streams to enable different types of information to flow across the network, so that both low-level details and high-level semantics can be utilized. Specifically, the feature maps from the last residual block of each stage of the backbone are fed into MSB and their channels are unified to 256 by a $3\times3$ convolutional layer. Since the boundary features interact with the mainstream, thus they convey semantic-related information. To make the boundary regions salient, the features of multi-scales are directly supervised by the binary boundary labels generated from the segmentation ground-truth. Here, a $1\times 1$ convolutional layer and a Sigmoid function is used to map the boundary features to edge maps, and all scales of boundary features are resized to 1/8 size of the input image and they are concatenated together for the following context aggregation. 

\vspace{-0.3cm}
\subsection{Boundary guided Context Aggregation (BCA) module}

We exploit the intrinsic partitioning capability of boundaries and use them as the harvested contexts to reinforce intra-class consistency. Inspired by the success of Non-local module in modeling long-range dependency in semantic segmentation, an attention-based module is developed, \textit{i.e.,} Boundary guided Context Aggregation module (BCA), to aggregate the context semantics. Unlike the original Non-local module that calculates self-attention, the boundary features serve as the \emph{Key} in the proposed BCA modules, where multi-scale features with rich semantics are used and only the boundaries regions are salient. As such, pixels from the same object activate similar attention areas, while that from different objects share less similarities. Figure \ref{fig2} shows the structure of the proposed BCA module. Specifically, given a semantic feature map $A \in \mathbb{R}^{C_1 \times H \times W}$ generated from the backbone and a boundary feature map $B \in \mathbb{R}^{C_2 \times H \times W}$ from the MSB, they are processed by two convolutional layers to generate two new feature maps $\{A_1,B_1\} \in \mathbb{R}^{C \times H \times W}$, where $C=256$. The features are then reshaped to $\mathbb{R}^{C \times N}$, where $N=H \times W$ is the number of pixels. We conduct matrix multiplication between the transpose of reshaped $A_1$ and $B_1$ and then apply a Softmax function. The whole operation can be described as:

\vspace{-0.2cm}
\begin{equation}\label{1}
F(i,j) = \frac{exp(B_{1i}\cdot {A_{1j}}^T)}{\sum^{N}_{i=1}{exp(B_{1i}\cdot {A_{1j}}^T)}}
\end{equation}
where $F$ is the boundary-semantic similarity map and $F(i,j)$ indicates the effect of $i^{th}$ position in the boundary feature map $B$ on the $j^{th}$ position in the semantic feature map $A$. 

The semantic context along the boundaries in $A$ will be aggregated into inner pixels. We therefore process $A$ with two convolutional layers of kernel size $1\times 1$ and reshape the output $A_2$ to  $\mathbb{R}^{C \times N}$, then perform matrix multiplication between $A_2$ and the boundary-semantic similarity map $F$, followed by an element-wise sum operation with feature $A$. For the $j^{th}$ pixel, its new feature can be calculated by:
\vspace{-0.2cm}
\begin{equation}\label{2}
D_j = A_j + \sum^{N}_{i=1}{F(i,j)\cdot A_{2i}}
\vspace{-0.2cm}
\end{equation}
where $D$ is the enhanced feature map after boundary context aggregation. Since the positions in the boundary regions of the same category will be activated with much higher weights than the other irrelevant ones, thus Equation (2) can be approximately written as:
\vspace{-0.2cm}
\begin{equation}\label{3}
D_j = A_j + \sum_{ i\in Boundary}{F(i,j)\cdot A_{2i}}
\end{equation}
which indicates that intra-class pixels achieve similar updates.  Compared with the previous Non-local module based methods that directly model the dependency between pixels, boundary information can better guarantee the intra-class consistency when ambiguity occurs.
\vspace{-0.3cm}

\subsection{Training Loss}

The proposed method consists of two main components which aim to generate the segmentation masks and boundaries, respectively. For semantic segmentation, the standard cross-entropy $L_{seg}$ is exploited to assign every pixel a category label. For boundary prediction, the binary cross-entropy loss $L_{boundary}$ is applied on MSB module, written as: 
\begin{equation}\label{4}
L_{boundary} = -\sum_{i}{(b_i \log{\hat{b_i}}+(1-b_i)\log{(1-\hat{b_i})})}
\end{equation}
where $b_i$ and $\hat{b_i}$ are the ground-truth and predicted boundaries, respectively.
To enhance the consistency of the two collaborative tasks, we also utilize another boundary-aware loss, \textit{i.e.} $L_{att}$ \cite{takikawa2019gated,li2020improving}, to describe the segmentation accuracy of the pixels along the boundary regions:
\begin{equation}\label{5}
L_{att} = -\sum_{i}\sum_{c}\mathbbm{1}[\hat{b_i}>t]\cdot (s_{i,c}\log{\hat{s_{i,c}}})
\end{equation}
where $t$ is the threshold of positive boundary, $s_{i,c}$ and $\hat{s_{i,c}}$ are the ground-truth and segmentation result of the $i$-th pixel of class $c$, respectively. Inspired by \cite{zhao2017pyramid}, we further apply an auxiliary cross-entropy loss $L_{aux}$ to the intermediate feature representations of the backbone to accelerate model convergence. Finally, the system training loss can be written as:
\begin{equation}\label{6}
L = \lambda_1 L_{seg} + \lambda_2 L_{boundary} + \lambda_3 L_{att} + \lambda_4 L_{aux}
\end{equation}
where $\lambda_1,\lambda_2,\lambda_3$ and $\lambda_4$ are the hyper-parameters, we empirically set $\lambda_1=1,\lambda_2=20,\lambda_3=1$ and $\lambda_4=0.4$.

\vspace{-0.2cm}
\section{Experiments}

We conduct extensive experiments on the benchmark datasets of Cityscapes \cite{cordts2016cityscapes} and ADE20K \cite{zhou2017scene}, and make fair comparison with the state-of-the-art counterparts.

\subsection{Implementation Details}

The model is implemented in Pytorch. ResNet-101 is used as the backbone and the dilation strategy is applied as in \cite{chen2017deeplab}. After the BCA module, the network is tailed with two $3 \times 3$ convolutional layers and one $1 \times 1$ convolutional layer to generate the segmentation mask. The resolution of the features to generate the final segmentation result is 1/8 of the initial input and bilinear interpolation is used to upsample the prediction. Data augmentation is performed during training as in \cite{zhao2017pyramid}. The initial learning rate is set to 0.01, and momentum  is set to 0.9 with weight decay of 0.0005. We use the polynomial learning rate adjustment policy $\gamma = \gamma_0 \times (1-\frac{iter}{max_{iter}})^{power}$, where $\gamma$ and $\gamma_0$ are the current and initial learning rate, respectively, and power is set to $0.9$. The batch size is set to 8 for Cityscapes and 12 for ADE20K, and the models are trained for $200$ and 120 epochs on these two datasets, respectively. During inference, we follow the multi-scale inference strategy as in \cite{{zhang2018context},{zhao2017pyramid},{ding2019boundary}}. The open source code provided by \cite{takikawa2019gated} is utilized to generate the ground-truth boundaries from the segmentation masks and the boundary radius is set to 2.

\vspace{-0.1cm}
\subsection{Model Analysis}
\vspace{-0.2cm}
\textbf{Ablation study.} As shown in Table \ref{table1}, simply incorporating the MSB module into the network has little influence on the performance of the dilated FCN baseline model. When introducing the boundary guided context aggregation mechanism but with a single-scale boundary (SSB) extractor, the performance has significantly improved by 3.54\%, demonstrating that the proposed information aggregation manner can indeed aggregate more reasonable context. By utilizing the proposed MSB module to perform boundary detection, the segmentation performance continues to improve and 80.03\% mIoU has been achieved, which clearly validates the significance of the multi-scale design of the proposed MSB module. Besides, the boundary-aware loss leads to an additional mIoU improvement of 0.89\%, clearly validating its effectiveness.

\begin{table}[ht]
\setlength{\belowcaptionskip}{-0.2cm}
\setlength{\abovecaptionskip}{0.1cm}
\renewcommand\arraystretch{1.2}
\centering
\begin{tabular}{ccccc}
\hline
\textbf{MSB} & \textbf{SSB} &\textbf{BCA} &\textbf{$L_{att}$}  & \textbf{mIoU\%} \\ \hline
             &                                &             &                    &    75.52       \\ \hline
\checkmark   &                                &             &                    &    74.95       \\ \hline
             &  \checkmark                    &\checkmark   &                    &    79.06       \\ \hline
\checkmark   &                                &\checkmark   &                    &    80.03       \\ \hline
\checkmark   &                                &\checkmark   &   \checkmark       & \textbf{80.92} \\ \hline    
\end{tabular}
\caption{Ablation study on MSB and BCA modules and the boundary-aware training loss.}
\label{table1}
\end{table}

\textbf{Comparison of different context aggregation modules.} The results are shown in Table \ref{table2}. Specifically, ASPP \cite{chen2017deeplab} applies parallel convolutions to capture multi-scale context information, its potential in guiding the mainstream context aggregation is not fully exploited. Our model achieves 0.74\% accuracy improvement than ASPP. Compared to the previous self-attention based methods, such as Non Local \cite{wang2018non}, RCCA \cite{huang2019ccnet} and DNL \cite{yin2020disentangled}, BCA can effectively alleviate the adverse effect caused by intra-class ambiguity, leading to significant mIoU improvements of 1.43\%, 1.59\% and 0.65\%, respectively. Moreover, different from BFP \cite{ding2019boundary} that exploits boundaries as a constraint for feature propagation, the proposed BCA is not sensitive to the closure of boundaries and it achieves 4.15\% mIoU improvement.

\begin{table}[ht]

\setlength{\belowcaptionskip}{-0.2cm}
\setlength{\abovecaptionskip}{0.1cm}
\renewcommand\arraystretch{1.2}
\centering
\resizebox{\linewidth}{!}{
\begin{tabular}{l|ccccccc}\toprule[1pt]
\textbf{Metric} & \textbf{Dilated FCN} & \textbf{+ASPP\cite{chen2017deeplab}} & \textbf{+Non Local\cite{wang2018non}} & \textbf{+RCCA\cite{huang2019ccnet}} & \textbf{+DNL\cite{yin2020disentangled}} & \textbf{+MSB-BFP\cite{ding2019boundary}} & \textbf{+MSB-BCA }\\ \midrule[1pt]
mIoU\% &   75.52 & 79.29 & 78.60 & 78.44 &  79.38 & 75.88 & \textbf{80.03}  \\ \bottomrule[1pt]

\end{tabular}}
\caption{Comparison of context aggregation modules.}
\label{table2}
\end{table}

\textbf{Performance along the boundaries.} In Table \ref{table3}, we evaluate the segmentation accuracies in the boundary and interior regions, and compare the proposed BCANet with one state-of-the-art method, \textit{i.e.}, SegFix \cite{yuan2020segfix}, which is specially designed to refine the segmentation accuracy along the boundaries. Following \cite{yuan2020segfix}, we employ the boundary F-score with threshold of 0.0003 and the interior F-score to measure the performance. Compared to dilated FCN, both SegFix and BCANet improve the accuracies, while our method performs more favorably in the interior regions and SegFix works better along the boundaries, clearly validating the context aggregation ability of our method. Besides, when combing SegFix with our method, higher accuracy is reported which confirms the complementary of our method and SegFix.

\begin{table}[ht]
\setlength{\belowcaptionskip}{-0.2cm}
\setlength{\abovecaptionskip}{0.1cm}
\renewcommand\arraystretch{1.2}
\centering
\begin{tabular}{l|cccc}\toprule[1pt]
Metrics            &Dilated FCN & +SegFix    & +MSB+BCA & +MSB+BCA+SegFix \\ \midrule[1pt]
F-score (boundary)  &  57.93     &   61.70    &      60.38    &   \textbf{63.92}  \\ \hline
F-score (interior)   &  75.07     &   75.26   &      77.02    &   \textbf{77.13}  \\ \hline
mIoU              &  75.52     &   76.39    &      80.03    &   \textbf{80.86}       \\ \bottomrule[1pt]
\end{tabular}
\caption{Comparison of segmentation accuracy in the boundary and interior regions.}
\label{table3}
\vspace{-0.1cm}
\end{table}

\subsection{Visualization}

\begin{figure*}[ht]
\vspace{-0.3cm}
\centering
\subfigure[Image]{
\begin{minipage}[t]{0.18\linewidth}
\centering
\includegraphics[width=1\linewidth]{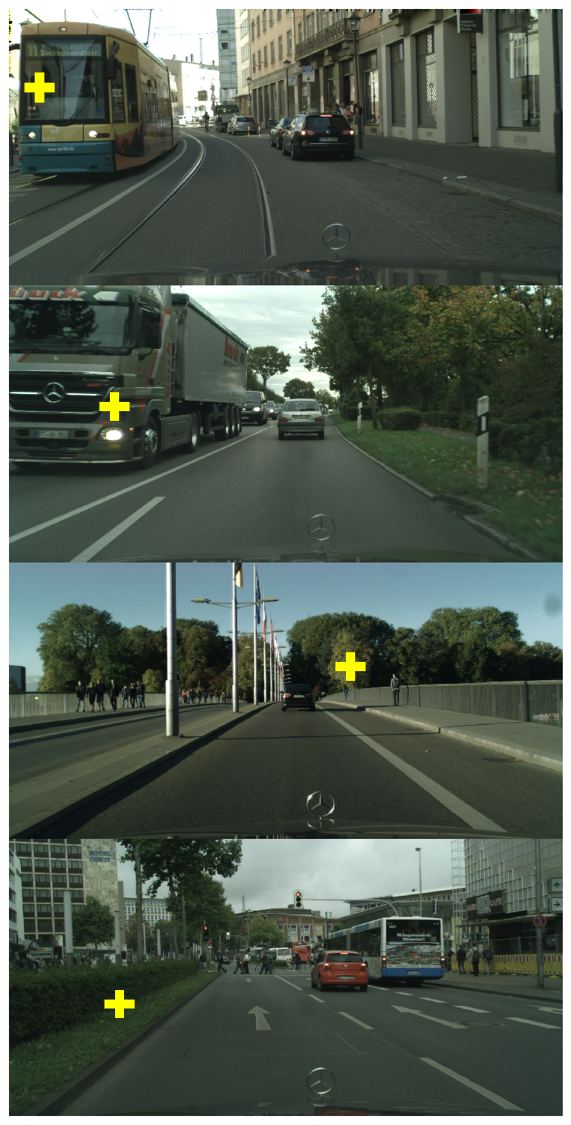}
\end{minipage}
}
\subfigure[ASPP]{
\begin{minipage}[t]{0.18\linewidth}
\centering
\includegraphics[width=1\linewidth]{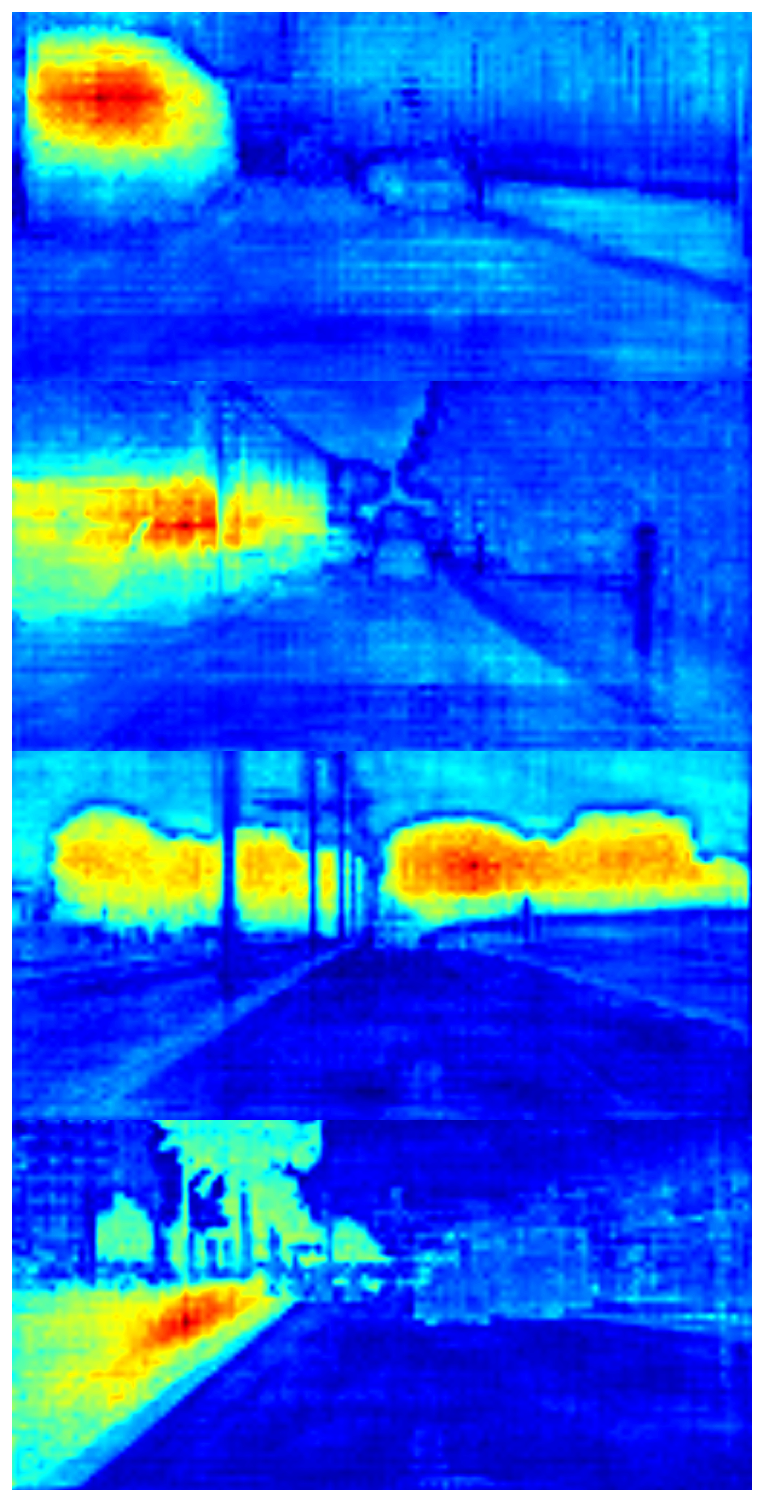}
\end{minipage}
}
\subfigure[RCCA]{
\begin{minipage}[t]{0.18\linewidth}
\centering
\includegraphics[width=1\linewidth]{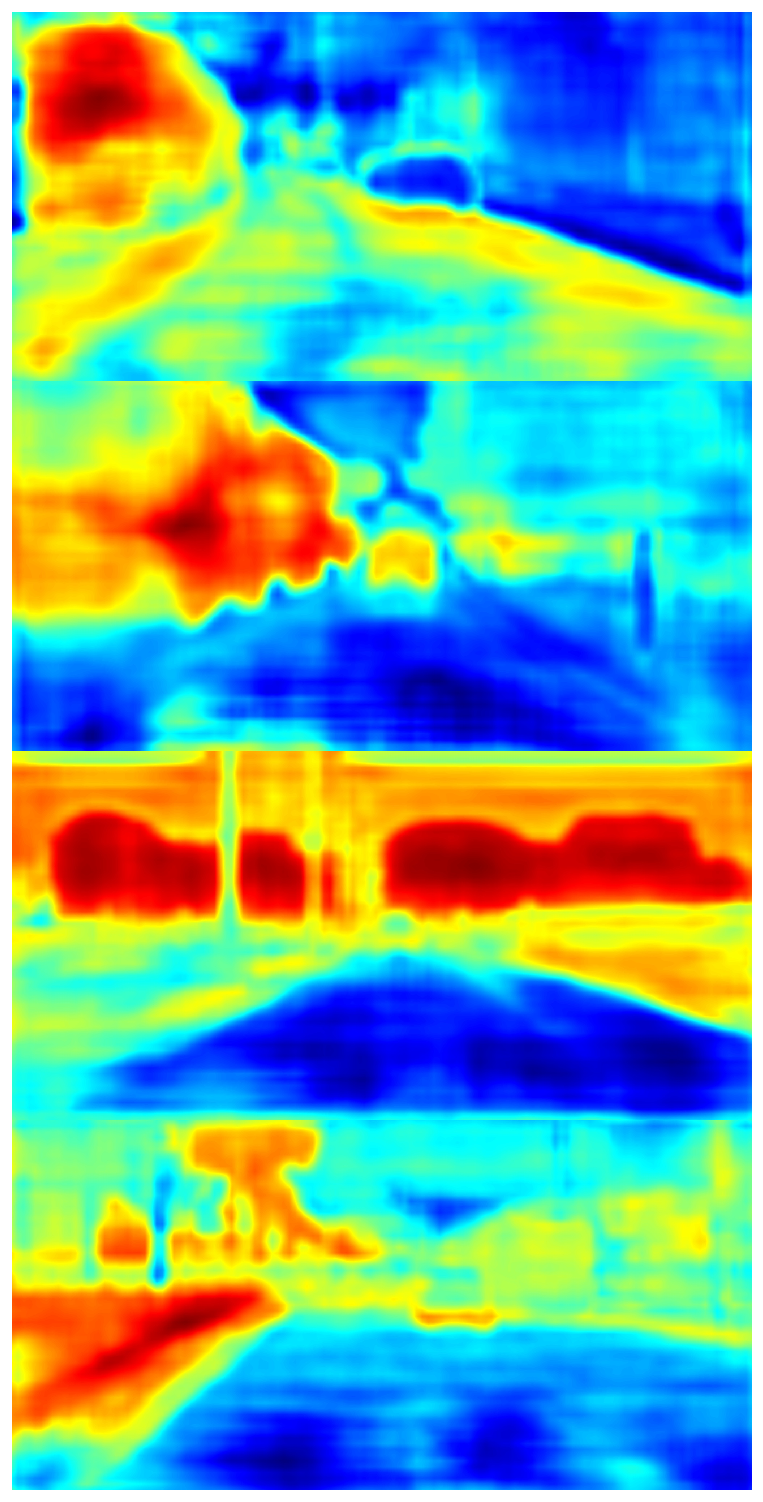}
\end{minipage}
}
\subfigure[BFP]{
\begin{minipage}[t]{0.18\linewidth}
\centering
\includegraphics[width=1\linewidth]{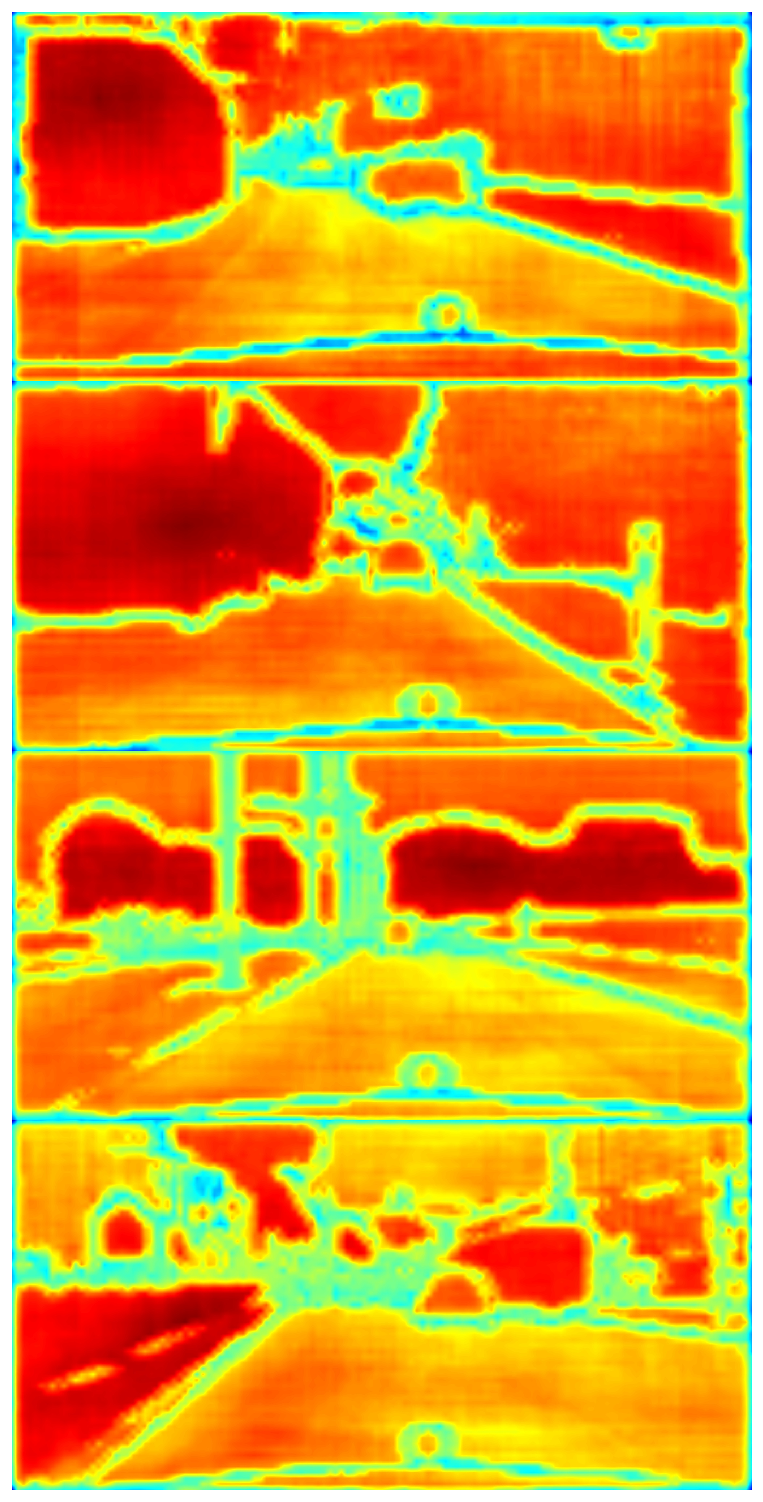}
\end{minipage}
}
\subfigure[Ours]{
\begin{minipage}[t]{0.18\linewidth}
\centering
\includegraphics[width=1\linewidth]{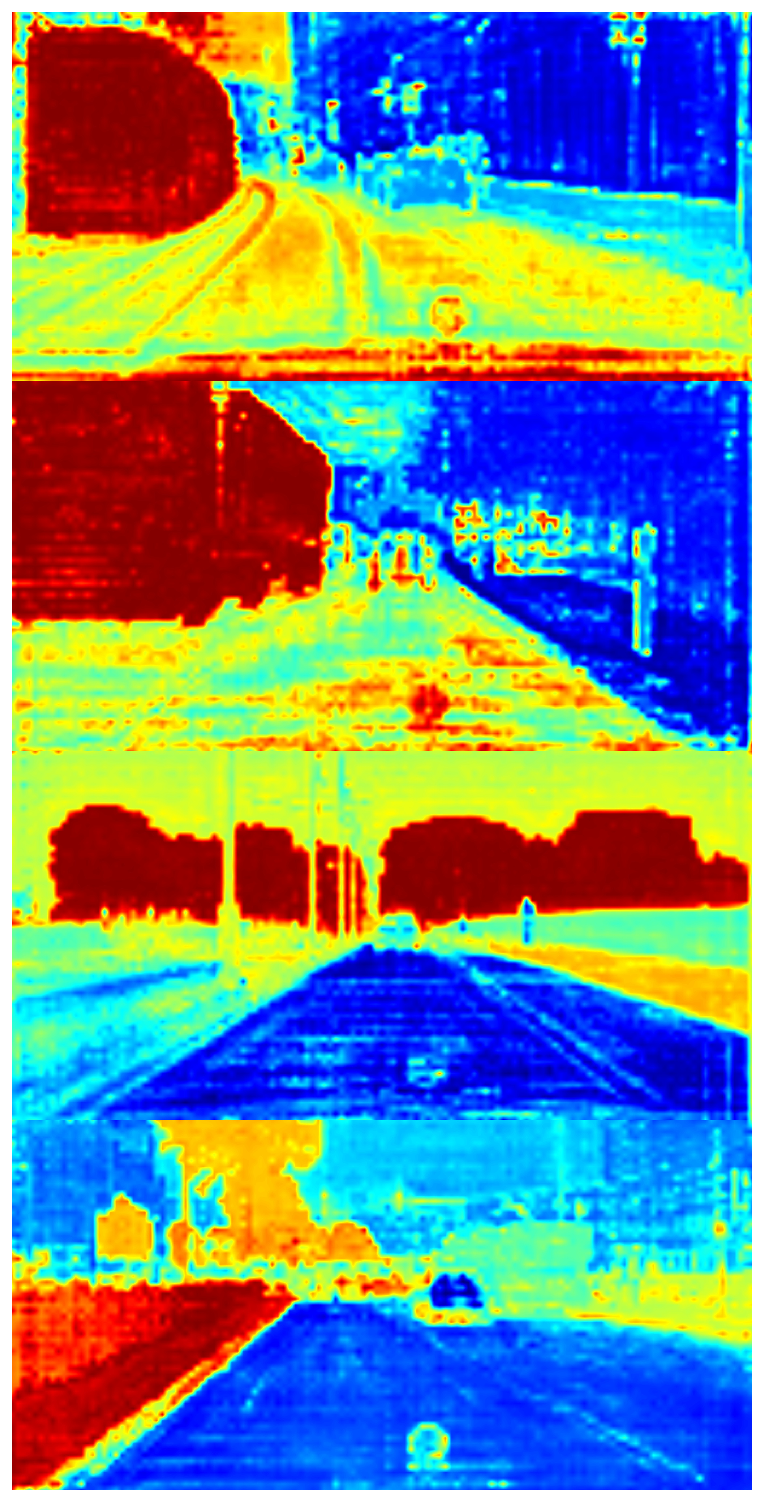}
\end{minipage}
}
\caption{Visualization of the cosine similarity between the marked pixel and the whole feature map after context aggregation. (a) The input image; (b) ASPP \cite{chen2017deeplab}; (c) RCCA \cite{huang2019ccnet}, which is an improved version of Non-local module; (d) BFP \cite{ding2019boundary}; and (e) the proposed BCA module.} 
\label{fig4}
\vspace{-0.3cm}
\end{figure*}

\textbf{Feature similarity visualization.} BCA is designed to make the pixels from the identical category achieve mutual gains. Accordingly, we visualize the feature similarity after context aggregation to see if the method is able to enhance the intra-class consistency and inter-class discrimination. We use cosine distance to measure the similarity between two features, and the results are visualized in Figure \ref{fig4}, where the reference pixel is marked by the yellow cross. In particular, ASPP \cite{chen2017deeplab} aggregates pixels of the same category efficiently, but the intra-class consistency tends to be weak for large objects, mainly because it neglects the long-range context. The RCCA module \cite{huang2019ccnet} in CCNet is an improved version of Non-local module which aggregates the full-image contexts. In this counterpart, the pixels from the same category indeed have larger similarities; however, it also introduces many unreliable contexts, indicating that the Non-local based module cannot exclude inter-class dependencies. In terms of BFP \cite{ding2019boundary}, it can be obviously observed that inter-class discrimination is weak that the context information will flow across the whole image when the boundary closure cannot be guaranteed. Comparatively, our method performs more favorably, demonstrating that boundary context is effective for semantic segmentation.

\vspace{0.2cm}
\begin{figure*}[ht]
\centering
\subfigure[]{
\begin{minipage}[t]{0.22\linewidth}
\centering
\includegraphics[width=1\linewidth]{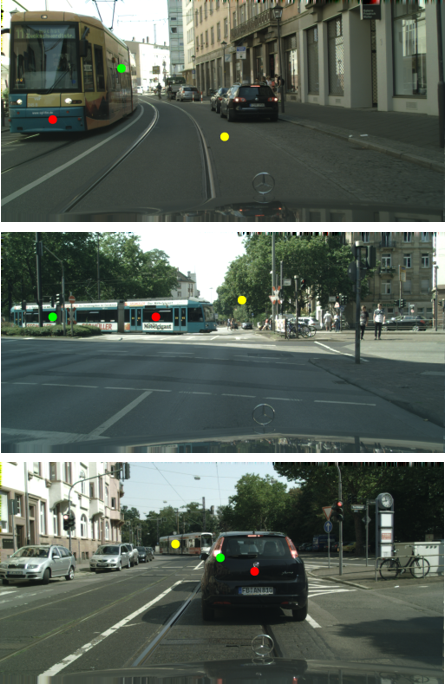}
\end{minipage}
}
\subfigure[]{
\begin{minipage}[t]{0.22\linewidth}
\centering
\includegraphics[width=1\linewidth]{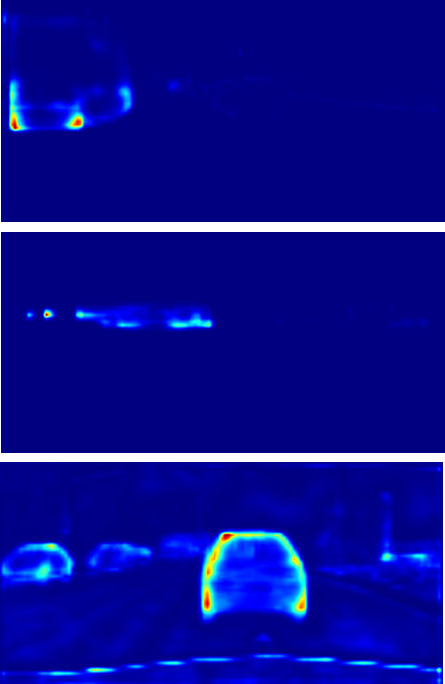}
\end{minipage}
}
\subfigure[]{
\begin{minipage}[t]{0.22\linewidth}
\centering
\includegraphics[width=1\linewidth]{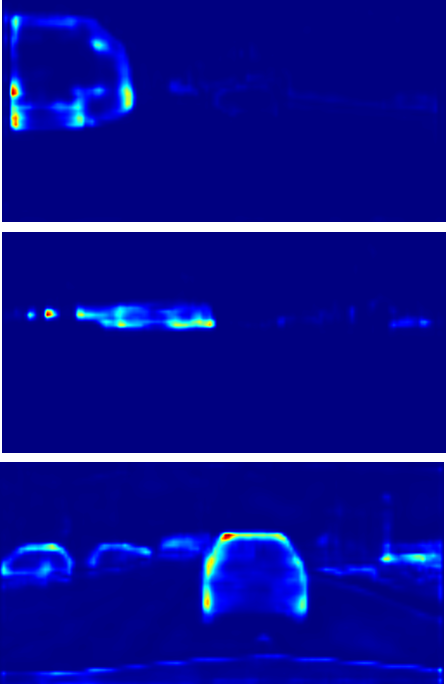}
\end{minipage}
}
\subfigure[]{
\begin{minipage}[t]{0.22\linewidth}
\centering
\includegraphics[width=1\linewidth]{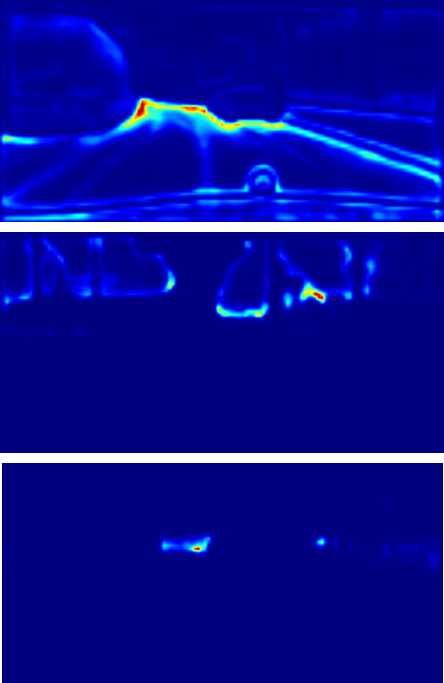}
\end{minipage}
}
\caption{Visualization of the attention maps. (a) The input image; attention maps of (b) the red point, (c) the green point, and (d) the yellow point. } 
\label{fig5}
\vspace{-0.4cm}
\end{figure*}

\vspace{-0.3cm}
\textbf{Attention map visualization.} In the proposed BCA module, when aggregating the contexts into a marked reference point, positions along the corresponding boundary regions are expected to be activated with much higher weights. To validate this assumption, we visualize the attention map of different positions in Figure \ref{fig5}. In particular, the red and green points belong to the same class and they have similar attention maps, but the attention map of yellow point is significantly different. The visualization results show that the proposed method indeed assists to enhance the intra-class consistency and inter-class discrimination.

\begin{figure*}[ht]
\centering
\subfigure[Image]{
\begin{minipage}[t]{0.18\textwidth}
\centering
\includegraphics[width=1\linewidth]{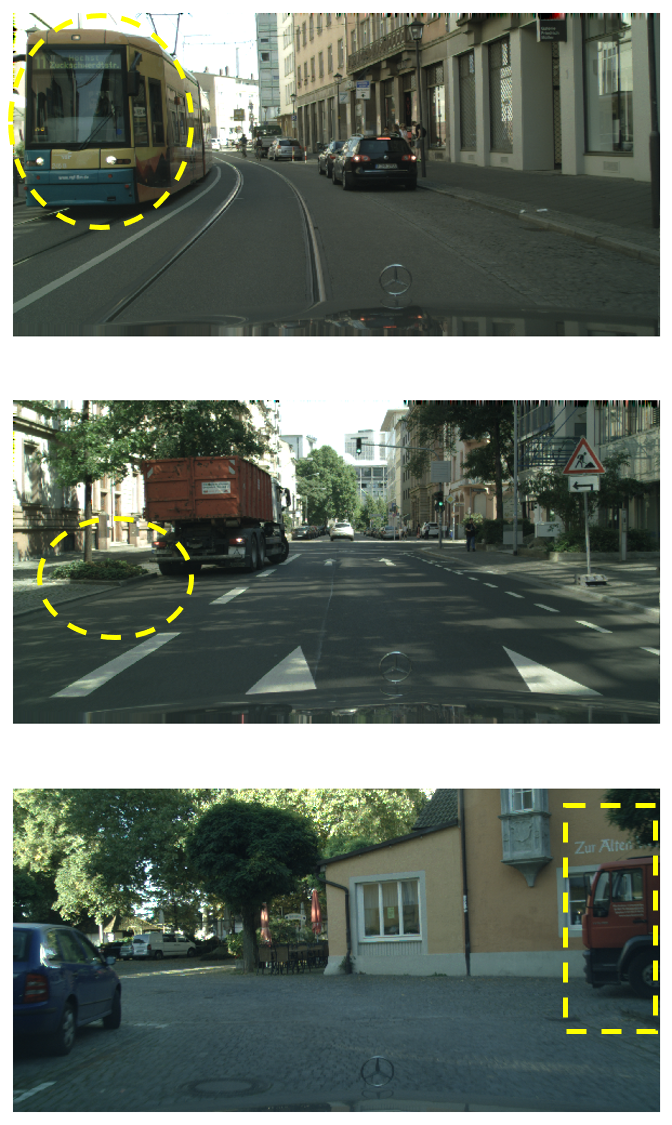}
\end{minipage}
}
\subfigure[GT]{
\begin{minipage}[t]{0.18\textwidth}
\centering
\includegraphics[width=1\linewidth]{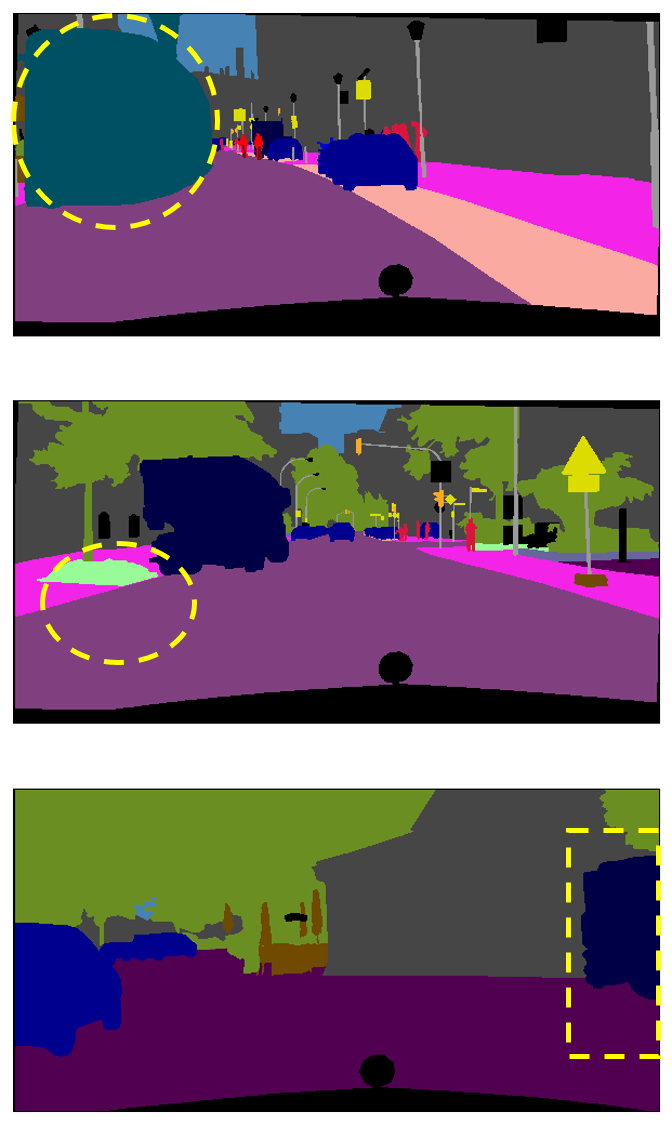}
\end{minipage}
}
\subfigure[CCNet]{
\begin{minipage}[t]{0.18\textwidth}
\centering
\includegraphics[width=1\linewidth]{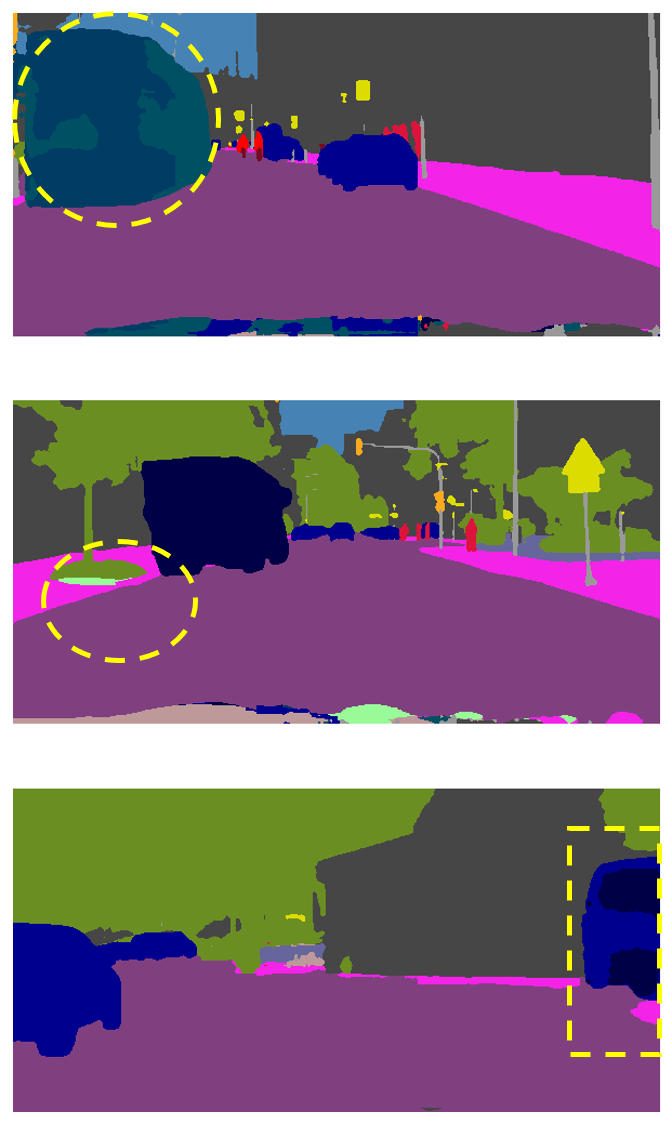}
\end{minipage}
}
\subfigure[BFP]{
\begin{minipage}[t]{0.18\textwidth}
\centering
\includegraphics[width=1\linewidth]{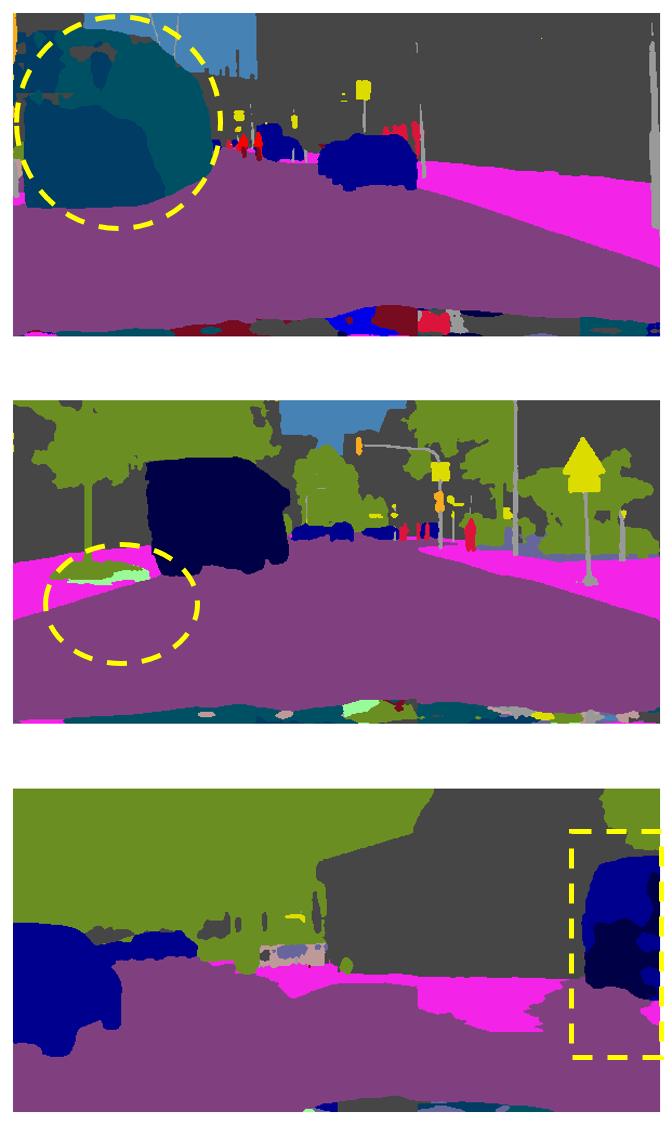}
\end{minipage}
}
\subfigure[Ours]{
\begin{minipage}[t]{0.18\textwidth}
\centering
\includegraphics[width=1\linewidth]{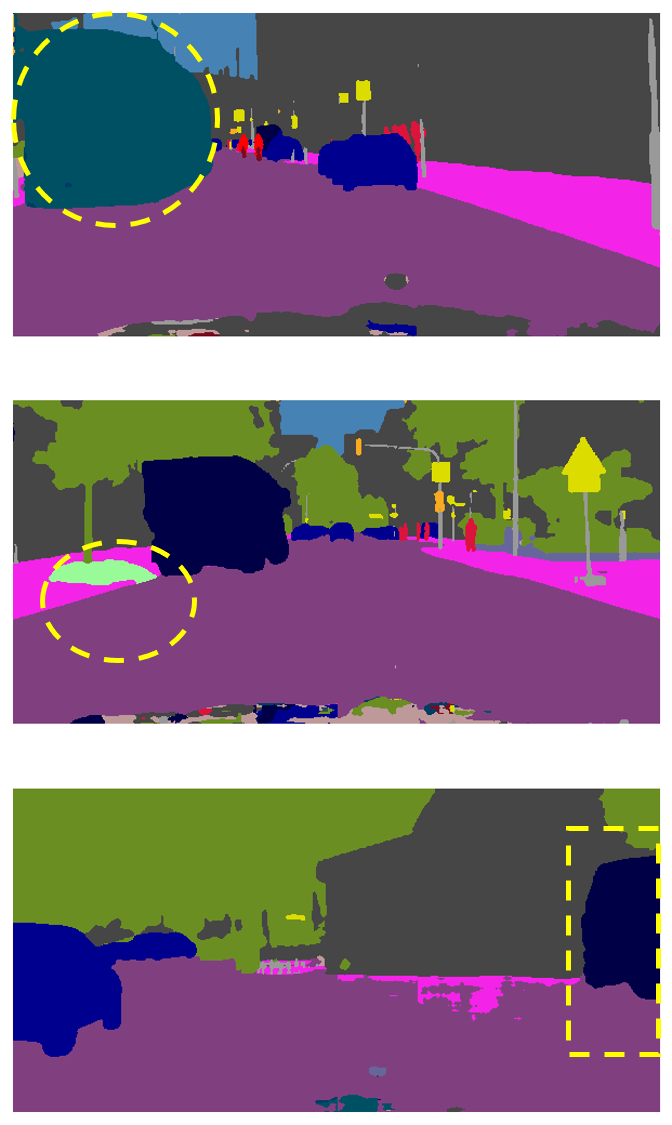}
\end{minipage}
}
\caption{Comparison of the final segmentation results. (a) The initial image; (b) the ground-truth; (c) CCNet \cite{huang2019ccnet}; (d) BFP \cite{ding2019boundary}; (e) the proposed BCANet.} 
\label{fig6}
\vspace{-0.4cm}
\end{figure*}

\textbf{Segmentation result visualization.} Figure \ref{fig6} shows the final segmentation results achieved by BCANet and the other methods modeling long-range contexts. Take the images in the first row for example, BCANet successfully segments the complete bus while the compared counterparts are struggling with the complex interior variations. In second row, the proposed model is able to separate the ground and the tree more clearly and it segments the truck more accurately in the third row. All the results have validated the effectiveness of our method.
\vspace{-0.3cm}


\begin{table}[ht]
\centering
\begin{tabular}{l|c|c|c|c}
\hline
\multicolumn{2}{l|}{~}&\multicolumn{1}{|c|}{\textbf{Cityscapes}} & \multicolumn{2}{|c}{\textbf{ADE20K}} \\ \hline
\textbf{Model} & \textbf{Backbone}    & \textbf{mIoU\%} & \textbf{mIoU\%} & \textbf{pixAcc\%}\\ \hline
RefineNet\cite{lin2017refinenet}      &ResNet-101    &73.6    &40.2  &-                                \\
GCN\cite{peng2017large}            &ResNet-101    &76.9    &-     &-                            \\
PSPNet\cite{zhao2017pyramid}         &ResNet-101    &78.4    &43.29 &81.39                                   \\
BiSeNet\cite{yu2018bisenet}        &ResNet-101    &78.9    &-     &-                                   \\
DFN\cite{yu2018learning}            &ResNet-101    &79.3    &-     &-                                   \\
PSANet\cite{zhao2018psanet}         &ResNet-101    &80.1    &43.77 &81.51                                   \\
EncNet\cite{zhang2018context}         &ResNet-101    &-      &44.65  &81.69                                   \\
ANL\cite{zhu2019asymmetric}            &ResNet-101    &81.3    &45.24 &-                                   \\
DANet\cite{fu2019dual}          &ResNet-101    &81.5    &-     &-                                   \\
CCNet\cite{huang2019ccnet}          &ResNet-101    &81.4   &45.22  &-                                   \\ 
GSCNN\cite{takikawa2019gated}          &WideResNet38  &\textbf{82.8}   &-      &-                                   \\
BFP\cite{ding2019boundary}            &ResNet-101    &81.4   &-      &-                                   \\
OCR\cite{yuan2020object}           &ResNet-101     &81.8   &45.28  &-        \\ 
RPCNet\cite{zhen2020joint}         &ResNet-101    &81.8   &-      &-                                   \\
CPNet\cite{yu2020context}          &ResNet-101    &81.3   &\textbf{46.27}  &81.85    \\ \hline
\textbf{BCANet}&ResNet-101    &81.7   &45.62  &\textbf{82.35}    \\ \hline
\end{tabular}
\vspace{0.3cm}
\caption{Segmentation results on Cityscapes test set and ADE20K validation set.}
\label{table4}
\vspace{-0.7cm}
\end{table}

\vspace{-0.5cm}
\subsection{Comparison with the State-of-the-art}

\textbf{On Cityscapes Test Set.} We conduct experiments on Cityscapes test set and quantitatively evaluate the proposed method on the official evaluation server. To ensure the comparison fairness, multi-scale inference and flipping strategies are exploited during testing. The compared methods include \cite{lin2017refinenet,peng2017large,zhao2017pyramid,yu2018bisenet,yu2018learning,zhao2018psanet,zhu2019asymmetric,fu2019dual,yuan2020object,huang2019ccnet,takikawa2019gated,ding2019boundary,zhen2020joint,yu2020context} and Table \ref{table4} shows the results. As can be seen, the proposed BCANet achieves 81.7\% mIoU on Cityscapes test set, which outperforms most of the existing studies. GSCNN \cite{takikawa2019gated} achieves the highest segmentation accuracy mainly because a stronger backbone and external dataset are utilized. RPCNet \cite{zhen2020joint} has cascaded more ResNet blocks to obtain the features of a larger resolution (1/4 initial input) to generate final results, which tends to benefit small objects, while our method and the other studies generally use the features of 1/8 scale of the given image. Among the models with fair settings, our results are comparable with the current state-of-the-art.

\textbf{On ADE20K Validation Set.} We compare BCANet with \cite{lin2017refinenet,zhao2017pyramid,zhao2018psanet,zhang2018context,huang2019ccnet,yuan2020object,zhu2019asymmetric,yu2020context}. As shown in Table \ref{table4}, the proposed method achieves the best performance of 82.35\% under the metric of pixel accuracy (pixAcc). In terms of mIoU, it achieves comparable result with most state-of-the-art methods while is a little inferior to CPNet \cite{yu2020context}. It is mainly because both intra-class context prior and the reversed prior are considered in CPNet, which favors the performance of all the classes. Comparatively, our model is more conducive to the situation that ambiguity occurs inside the object with relatively large spatial spans, thus it performs more favorably in terms of pixel accuracy.

\vspace{-0.5cm}
\section{Conclusion}
\vspace{-0.3cm}
In this paper, we propose a novel approach to semantic segmentation, which explicitly exploits boundary as the aggregated context to improve intra-class consistency. In particular, MSB and BCA modules are specially designed to perform such boundary guided context aggregation. On the two widely applied benchmarks of Cityscapes and ADE20K, we have experimentally demonstrated that the proposed model performs more favorably than the existing attention-based methods by leveraging the partitioning capability of boundary. It should be noted that it may theoretically incur errors when a dominating foreground object shares large boundary areas with the background, since its features tend to leak into the background. We observe the datasets and this case rarely happens, but we keep it in mind and will work for such a corner case in the future.

\vspace{-0.4cm}
\section*{Acknowledgements}
\vspace{-0.2cm}
This work is partly supported by the National Natural Science Foundation of China (No. 62022011, U20B2069), the Research Program of State Key Laboratory of Software Development Environment (SKLSDE-2021ZX-04), and the Fundamental Research Funds for the Central Universities.

\bibliography{egbib}
\end{document}